\documentclass{article}
\usepackage{spconf,amsmath,amssymb,graphicx}
\usepackage{cite}
\usepackage{svg}
\usepackage[hidelinks]{hyperref}

\title{Floorplan-Context-Aware Cross-Environment Human Activity Recognition from mmWave Radar Point Clouds}

\name{Zixiong Han, Xilai Wang, Chenzhe Zhao, Amr Mourad, and Miodrag Bolic}
\address{Computational Analysis and Applications Research Group,\\
University of Ottawa, Ottawa, Canada}

\begin{document}
\maketitle

\begin{abstract}
Human activity recognition (HAR) models based on mmWave radar point clouds and trained in controlled open spaces often experience performance degradation when deployed in furnished rooms. We propose a Floorplan-Context-Aware (FCA) HAR enhancement method that augments the input with subject--furniture relative position information to improve cross-environment performance. Because existing mmWave point-cloud datasets lack registered floorplans, we introduce a pseudo-landmark augmentation strategy for training the FCA model. In cross-environment evaluation on the mm-ADL-Home dataset, FCA achieves 77.70\% accuracy and 65.49\% macro-F1, improving upon the point-cloud-only direct-transfer baseline by 10.73 and 10.05 percentage points, respectively, and exceeding the strongest evaluated adaptation baseline by 7.37 and 6.27 percentage points. At deployment, FCA requires only a one-time registration of furniture landmarks, with no target-domain activity annotations or model weight updates.
\end{abstract}

\begin{keywords}
mmWave radar, human activity recognition, point clouds, floorplan context, cross-environment recognition
\end{keywords}

\section{Introduction}
\label{sec:introduction}
Millimeter-wave (mmWave) radar enables contact-free human activity recognition (HAR) without relying on illumination or RGB imaging, supporting smart homes and ambient assisted living \cite{amin2019radar, yu2022noninvasive}. Radar point clouds compactly represent human geometry and radial velocity for spatiotemporal recognition \cite{cui2024milipoint, wang2021m, yang2023mm, yu2022noninvasive}. However, models trained in controlled spaces often degrade in furnished rooms: multipath distorts point clusters, furniture occludes the radar line of sight, and room layouts affect movement and posture transitions.

These environmental effects have motivated both signal-processing and learning-based solutions \cite{chen2023environment, gong2025seradar, gorji2021generalization}. For systems that observe only sparse point clouds, transfer learning is the most practical paradigm because it can operate on a point-cloud encoder without raw radar measurements or high-resolution micro-Doppler spectrograms. Prior work studies active adaptation \cite{lin2026active}, unsupervised adaptation \cite{el2025radar, liu2024source}, and supervised transfer \cite{liu2022mtranssee}. However, most adaptation techniques were designed for 2D spectrograms, and their effectiveness on sparse mmWave point clouds remains under-explored and unstable.

We investigate an additional cue: the subject's position relative to \emph{landmark furniture} (furniture relevant to the activities). Radar point clouds are not only local kinematic imagers of human shape and motion but also global metric spatial locators. We therefore leverage both capabilities to exploit this cue. Postural actions are strongly associated with such landmarks: sitting and seated transitions (e.g., stand-to-sit and sit-to-stand) often occur at \emph{sittable} furniture (chairs/sofas), whereas lying and resting often occur at \emph{lieable} furniture (beds/couches). Walking is less tied to these landmarks, and falls can occur throughout the room. 

Incorporating spatial environment representations provides critical scene context for learning tasks \cite{kitani2012activity, gao2020vectornet, fan2020home}. Inspired by RF-Diary's person-centric floorplan encoding \cite{fan2020home}, we represent spatial context through subject--furniture proximity while further simplifying the representation by using proximity as a heuristic to identify the furniture the subject is most likely to interact with, rather than encoding the subject's spatial relationships with all mapped objects. Specifically, we extend a PointNet--LSTM point-cloud HAR backbone \cite{wang2026beyond} with a lightweight encoding of the nearest furniture landmark's category and distance. Frame-level point-cloud features and spatial-context features are fused before temporal modeling, enabling the model to learn point-cloud dynamics and changes in proximity to furniture landmarks. Because existing point-cloud HAR datasets do not provide floorplans, we also propose a source-domain pseudo-landmark augmentation to enable training without registered source maps. At deployment, the method requires only one-time registration of target landmark furniture, with no target activity annotations or weight updates.

On mm-ADL-Home, a cross-environment mmWave point-cloud HAR dataset comprising 14 participant--room settings across four furnished layouts, our FCA enhancement achieves 77.70\% accuracy and 65.49\% macro-F1, improving the original model by 10.73\% and 10.05\%, and exceeding the strongest evaluated adaptation baseline by 7.37\% and 6.27\%, respectively. These results show that floorplan context is effective when activity classes are physically correlated with indoor landmarks.

The main contributions of this work are:
\begin{itemize}
    \item A Floorplan-Context-Aware HAR enhancement strategy that augments a point-cloud-only HAR backbone with a lightweight floorplan-context encoder. The encoder captures furniture semantics and subject--furniture proximity and fuses them with the point-cloud representation for temporal modeling.
    \item A source-domain pseudo-landmark augmentation scheme that enables training the Floorplan-Context-Aware model using source point-cloud data without floorplan information.
\end{itemize}

\section{Methodology}
\label{sec:methodology}

\subsection{Furniture Registration and Frame-Level Context}
\label{subsec:presetup}
When deploying the HAR system in a new indoor environment, we perform a one-time setup to register landmark furniture in the radar coordinate system on the horizontal ($x$--$y$) plane. Coordinates can be measured with a radar-centric furniture coordinate registration system and stored with category IDs (e.g., 1 for \emph{sittable} Chair/Sofa and 2 for \emph{lieable} Bed/Couch). Both categories support sitting-related transition activities, whereas only the latter supports lying-related activities. Registered coordinates remain fixed during monitoring.

\label{subsec:tracking}

\begin{figure}[h]
    \centering
    \includegraphics[width=0.75\columnwidth]{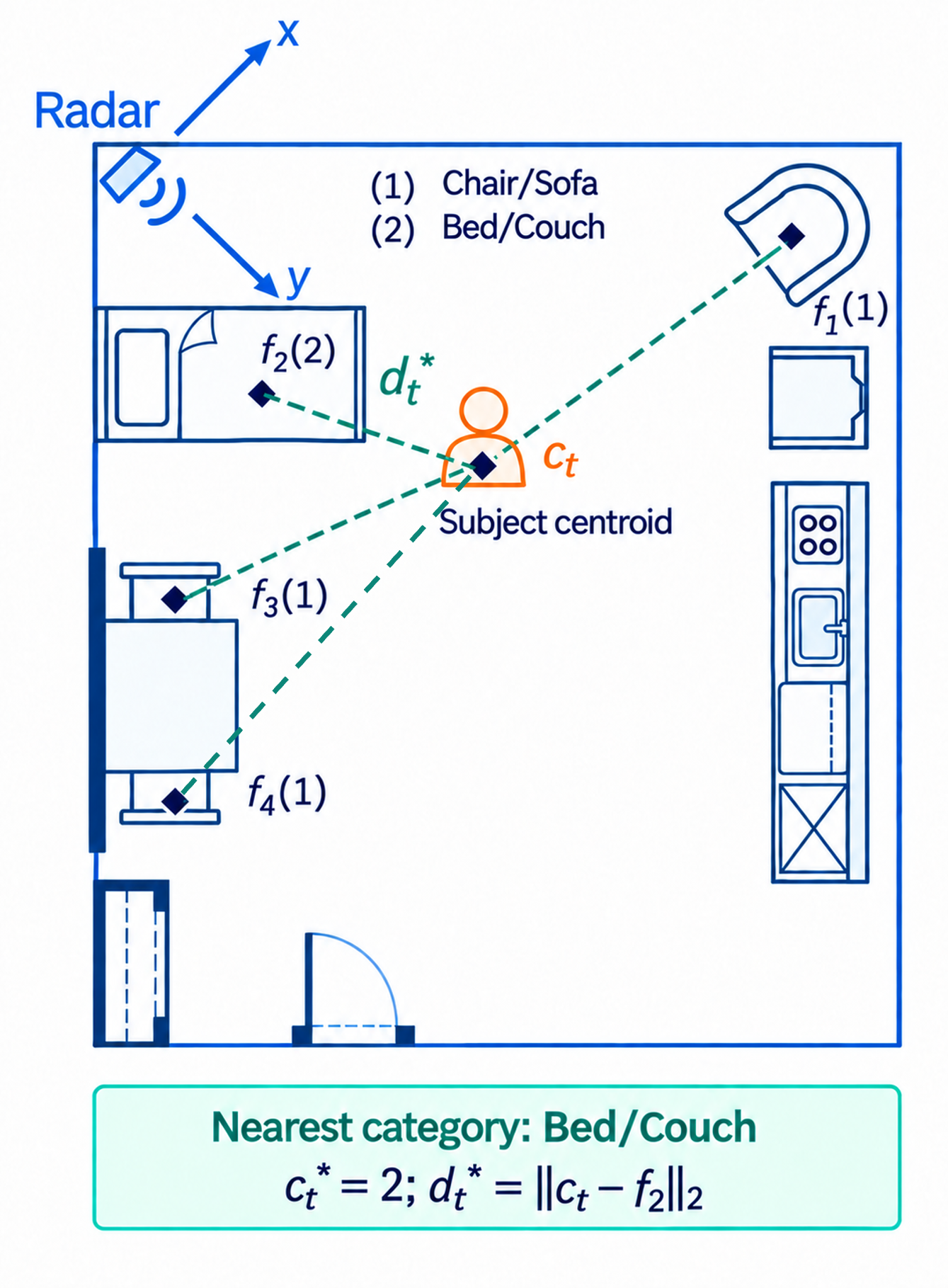}
    \caption{Illustrative spatial context from registered landmark furniture. The subject's horizontal centroid $\boldsymbol{c}_t$ determines the nearest-landmark category $c_t^*$ and distance $d_t^*$. $f_k$ denotes the coordinates of the $k$-th registered landmark, and the number in parentheses indicates its furniture type. In this example, the nearest landmark is Bed/Couch (ID 2).}
    \label{fig:spatial_context}
\end{figure}

An Extended Kalman Filter (EKF) group tracker with spatial gating tracks the subject, and the tracked point cloud is used to compute the subject centroid. Since the radar's moving target indicator (MTI) processing can suppress stationary returns, point clouds may be missing when the subject remains still. We therefore resample and interpolate the stream to a fixed rate of $T_f$ frames per second. For missing stationary frames, we reuse the most recent available point cloud of the tracked subject to preserve its centroid and set the radial velocity of every point to zero to represent the stationary state. For each synchronized frame, the horizontal centroid is
\begin{equation}
\mathbf{c}_t = (\bar{x}_t,\bar{y}_t)
=
\left(
\frac{1}{N}\sum_{i=1}^{N}x_{t,i},
\frac{1}{N}\sum_{i=1}^{N}y_{t,i}
\right),
\end{equation}
where $N$ is the number of points, $t$ denotes the frame index, and $i$ denotes the point index. The distance to the nearest landmark is
\begin{equation}
d_t^* = \min_k \left\| \mathbf{c}_t-\mathbf{f}_k \right\|_2,
\end{equation}
where $\mathbf{f}_k \in \{\mathbf{f}_1,\dots,\mathbf{f}_K\}$ is the coordinate of the $k$-th registered landmark and $K$ is the number of registered landmarks. The nearest-landmark category $c_t^*\in\{1,2\}$ and distance $d_t^*$ are appended to each point as $[x,y,z,v,c_t^*,d_t^*]$ to form a floorplan-context-augmented point cloud, as shown in Fig.~\ref{fig:spatial_context}.

\subsection{Model Architecture and Training}
\label{subsec:architecture}

We use point-cloud clips with $T$ consecutive frames to model temporal dynamics. The Floorplan-Context-Aware HAR model takes an input tensor of size $(T,N,6)$, where $N$ is the number of points per frame. After computing the context $[c_t^*,d_t^*]$, we normalize each frame by centering it at its horizontal centroid:
$\tilde{x}_{t,i}=x_{t,i}-\bar{x}_t$ and $\tilde{y}_{t,i}=y_{t,i}-\bar{y}_t$.
This removes absolute horizontal position from the point-cloud branch while keeping elevation $z$ and radial velocity $v$ unchanged.

\begin{figure*}[!t]
    \centering
    \includegraphics[
        width=\textwidth
    ]{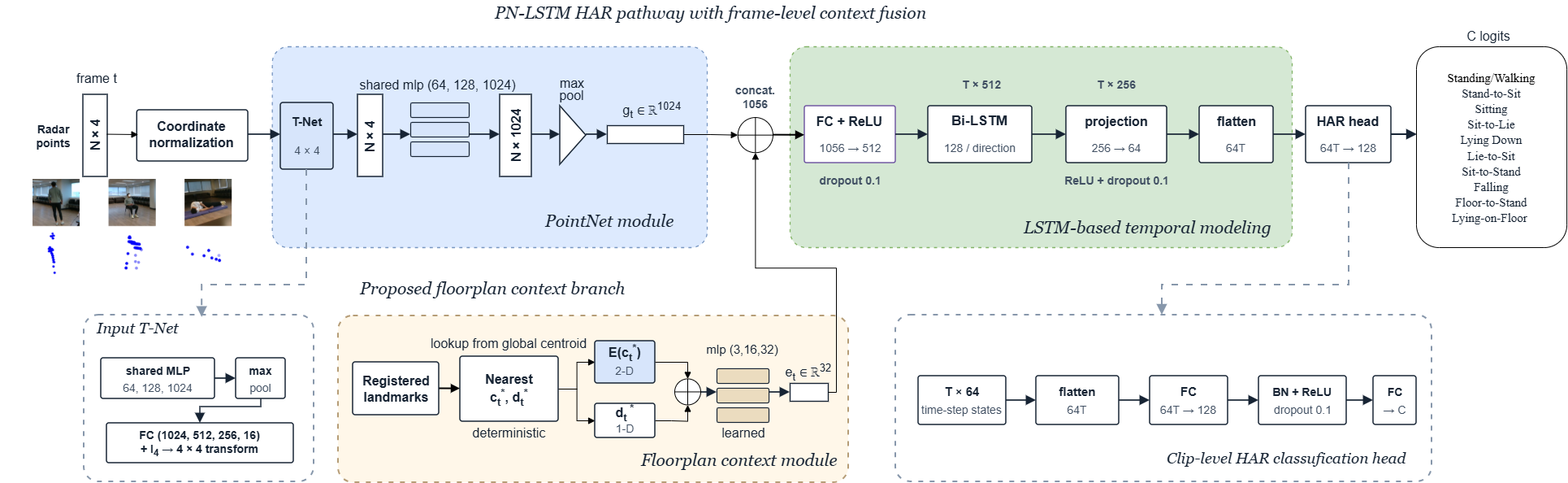}
    \caption{Architecture of the proposed Floorplan-Context-Aware HAR model. A framewise PointNet encodes normalized radar points, while the floorplan-context module encodes the category and distance of the nearest registered landmark, determined from the frame centroid in global radar coordinates. The 1024-D point-cloud feature and 32-D context feature are fused before Bi-LSTM temporal modeling and clip-level HAR classification.}
    \label{fig:fca_pnlstm}
\end{figure*}

The normalized point attributes $[\tilde{x},\tilde{y},z,v]$ and the context $[c_t^*,d_t^*]$ are processed by two branches: a PointNet module for point clouds and a context module for the nearest-landmark category and distance, as shown in Fig.~\ref{fig:fca_pnlstm}.


The PointNet module consists of an input T-Net, shared MLPs
($4\to64\to128\to1024$), and max-pooling, producing a frame-level point-cloud feature
$\mathbf{g}_t\in\mathbb{R}^{1024}$.
In the floorplan context module, a learned 2-D embedding of $c_t^*$ is concatenated with the scalar distance $d_t^*$ and passed through an MLP
($3\to16\to32$), producing
$\mathbf{e}_t\in\mathbb{R}^{32}$.
The two features are concatenated into
\[
\mathbf{u}_t
=
\mathbf{g}_t \oplus \mathbf{e}_t
\in\mathbb{R}^{1056}.
\]

The fused feature $\mathbf{u}_t$ is projected to
$\mathbf{z}_t\in\mathbb{R}^{512}$ using a fully connected layer, ReLU, and dropout ($p=0.1$). The projected frame features form the sequence
$\mathbf{Z}=(\mathbf{z}_1,\ldots,\mathbf{z}_T)
\in\mathbb{R}^{T\times512}$, which is processed by a single-layer Bi-LSTM with 128 hidden units per direction, producing a 256-D representation at each time step. These representations are projected from 256 to 64 dimensions using ReLU and dropout ($p=0.1$), forming
$\mathbf{H}\in\mathbb{R}^{T\times64}$. The sequence is then flattened into
$\mathbf{h}\in\mathbb{R}^{64T}$, corresponding to 2560 dimensions for $T=40$.

The classification head maps $\mathbf{h}$ through a fully connected layer
($64T\to128$), batch normalization, ReLU, dropout ($p=0.1$), and a final layer producing $C$ activity logits. The network is trained using cross-entropy loss:
\begin{equation}
\mathcal{L}_{\mathrm{CE}}
=
-\frac{1}{B}
\sum_{b=1}^{B}
\sum_{c=1}^{C}
y_{b,c}\log\hat{y}_{b,c},
\end{equation}
where $y_{b,c}$ is the one-hot label and $\hat{y}_{b,c}$ is the predicted softmax probability for class $c$. Adam is used with a learning rate of $10^{-3}$ and weight decay of $10^{-3}$.

\subsection{Spatial Context Augmentation in the Source Domain}
\label{subsec:augmentation}
Since existing mmWave point-cloud HAR datasets lack registered floorplans, we construct pseudo-landmarks from unnormalized source recordings with frame-level activity labels.

For recordings containing sitting and lying activities, we compute the mean centroid during \textit{sitting} and \textit{lying-down} periods and treat the resulting locations as Chair/Sofa and Bed/Couch anchors, respectively. We then assign the nearest landmark to each frame using the method described in Section~\ref{subsec:presetup}.

For recordings containing only locomotion activities (standing and walking), we place virtual anchors at typical room locations (e.g., 2.0--5.0\,m along the radar axis at intervals of 1--2\,m) to provide nearest-category and distance cues. For recordings containing simulated falls, we compute the mean centroid during the \textit{Lying-on-Floor} period and sample one virtual Chair/Sofa or Bed/Couch anchor at a distance
$d\sim\mathcal{U}(0.3,5.0)$\,m and angle
$\theta\sim\mathcal{U}(0,2\pi)$ relative to that centroid.

\begin{table*}[!t]
    \centering
    \caption{Cross-environment HAR evaluation on mm-ADL-Home.}
    \label{tab:mmadlhome_cross_environment}
    \renewcommand{\arraystretch}{1.10}
    \setlength{\tabcolsep}{3.0pt}
    \small
    \begin{tabular}{lcccccc}
        \hline
        Method & Acc. & Macro-F1 & Edit & F1@10 & F1@25 & F1@50 \\
        \hline
        Direct transfer & 66.97 $\pm$ 6.32 & 55.44 $\pm$ 7.52 & 49.64 $\pm$ 5.77 & 59.83 $\pm$ 6.02 & 56.76 $\pm$ 6.82 & 47.87 $\pm$ 7.90 \\
        DANN            & 69.01 $\pm$ 5.68 & 56.34 $\pm$ 8.22 & 49.48 $\pm$ 5.89 & 60.55 $\pm$ 6.02 & 57.79 $\pm$ 6.90 & 48.31 $\pm$ 8.19 \\
        Fine-tuning     & 69.36 $\pm$ 6.23 & 57.24 $\pm$ 7.48 & 50.90 $\pm$ 6.50 & 61.53 $\pm$ 6.17 & 58.80 $\pm$ 7.24 & 49.57 $\pm$ 7.87 \\
        CDAN            & 70.33 $\pm$ 7.13 & 59.22 $\pm$ 8.31 & 51.48 $\pm$ 6.11 & 62.74 $\pm$ 6.11 & 59.79 $\pm$ 7.28 & 51.36 $\pm$ 8.20 \\
        FCA-HAR & \textbf{77.70 $\pm$ 4.05} & \textbf{65.49 $\pm$ 4.95} & \textbf{60.70 $\pm$ 7.38} & \textbf{71.08 $\pm$ 5.61} & \textbf{69.59 $\pm$ 6.02} & \textbf{61.65 $\pm$ 7.42} \\
        \hline
    \end{tabular}
\end{table*}

\section{Experiments}
\label{sec:experiments}

\subsection{Datasets and Preprocessing}
\label{subsec:datasets}
We use mm-ADL \cite{wang2026beyond} as the source-domain dataset. It contains recordings from 11 subjects performing 10 activity classes: seven activities of daily living (ADLs) (standing/walking, stand-to-sit, sitting, sit-to-stand, sit-to-lie, lying-down, and lie-to-sit) and three simulated incident activities (falling, lying-on-floor, and floor-to-stand/recover). Point clouds are collected using a TI IWR6843ISK-ODS sensor in a controlled open-space environment.

For cross-environment evaluation, we collect mm-ADL-Home using the same sensor configuration and activity taxonomy. mm-ADL-Home contains 14 participant--room recordings across four furnished layouts; subjects perform ADLs freely to simulate daily life. Each recording lasts 15--20 minutes and includes at least 10 instances of each class. The dataset is publicly available at \url{https://github.com/OwenHan10/mm-ADL-Home.git}.

Both datasets are temporally resampled to $T_f=20$ Hz. Each point-cloud clip contains $T=40$ frames and $N=128$ points per frame. Frames containing more than 128 points are randomly downsampled, whereas frames containing fewer than 128 points are padded by resampling the available points \cite{wang2026beyond}. Window labels are assigned by majority voting over the frame-level labels. For source-domain training on mm-ADL, we construct a class-balanced training set comprising 50 point-cloud clips per activity per subject. For target-domain evaluation on mm-ADL-Home, point-cloud streams are segmented using a sliding window with a stride of 20 frames, corresponding to 50\% temporal overlap.

\subsection{Source-Domain Evaluation}
\label{subsec:source_loso}

We evaluate the models on mm-ADL using leave-one-subject-out (LOSO) cross-validation. The selected PointNet-LSTM backbone \cite{wang2026beyond} is compared with m-Activity \cite{wang2021m}, DVCNN \cite{yu2022noninvasive}, and the proposed Floorplan-Context-Aware model (FCA-HAR). FCA-HAR is trained on mm-ADL with pseudo-landmark augmentation applied independently to each subject's recording. Table~\ref{tab:mmadl_loso_har} summarizes the results. PointNet-LSTM achieves 86.0\% accuracy, outperforming DVCNN, the strongest prior baseline, by 6.7 percentage points. FCA-HAR achieves 89.8\% accuracy and 89.6\% macro-F1, exceeding PointNet-LSTM by 3.8 and 3.9 percentage points, respectively. These results demonstrate the benefit of incorporating floorplan context into the PointNet-LSTM backbone under the evaluated LOSO protocol.

\begin{table}[h]
    \centering
    \caption{Source evaluation performance on mm-ADL (\%).}
    \label{tab:mmadl_loso_har}
    \renewcommand{\arraystretch}{1.08}
    \setlength{\tabcolsep}{1.8pt}
    \small
    \begin{tabular}{lcccc}
        \hline
        Method & Acc. & Prec. & Recall & F1 \\
        \hline
        m-Activity  & 74.5$\pm$5.7 & 76.1$\pm$4.9 & 74.4$\pm$6.0 & 74.0$\pm$6.3 \\
        DVCNN       & 79.3$\pm$5.9 & 80.2$\pm$4.7 & 79.1$\pm$5.6 & 78.8$\pm$5.6 \\
        PointNet-LSTM     & 86.0$\pm$3.9 & 86.9$\pm$3.0 & 85.8$\pm$4.1 & 85.7$\pm$3.9 \\
        FCA-HAR & \textbf{89.8}$\pm$\textbf{3.0} &
                      \textbf{90.4}$\pm$\textbf{2.5} &
                      \textbf{89.6}$\pm$\textbf{3.1} &
                      \textbf{89.6}$\pm$\textbf{3.3} \\
        \hline
    \end{tabular}
\end{table}

\subsection{Target-Domain Cross-Environment Evaluation}
\label{subsec:eval_protocol}
We evaluate cross-environment deployment on mm-ADL-Home using PointNet-LSTM pretrained on mm-ADL, comparing direct transfer, the proposed FCA-HAR, unsupervised adaptation (DANN \cite{ganin2016domain} and CDAN \cite{long2018conditional}), and supervised target fine-tuning. The adaptation baselines follow prior radar cross-environment studies \cite{el2025radar, liu2022mtranssee}.

\textbf{Protocol.} Since adaptation requires target-domain data, for each participant--room recording, we use the first 20\% of chronological windows as a target-domain calibration set for training the adaptation baselines and reserve the remaining windows for testing. Each calibration set contains windows from all ten activity classes. We exclude a boundary margin to avoid overlap between calibration and test windows.

\textbf{Metrics.} We report window-level accuracy and macro-F1, the normalized segmental edit score computed from run-length-collapsed predicted and ground-truth label sequences, and segmental F1 scores at intersection-over-union thresholds of 10\%, 25\%, and 50\% \cite{lea2016temporal}. The temporal metrics assess activity ordering, fragmentation, and segment overlap beyond window-level classification.

\subsection{Cross-Environment Results}
\label{subsec:cross_env_results}

Table~\ref{tab:mmadlhome_cross_environment} summarizes the cross-environment results on mm-ADL-Home. Direct transfer achieves 66.97\% accuracy and 55.44\% macro-F1. Among the adaptation baselines evaluated with the specified calibration budget, CDAN achieves the highest accuracy and macro-F1 of 70.33\% and 59.22\%, respectively. Without target-room weight updates, FCA-HAR achieves the highest mean score on every reported metric, reaching 77.70\% accuracy and 65.49\% macro-F1 and exceeding CDAN by 7.37 and 6.27 percentage points, respectively. The edit score increases from 51.48 to 60.70, while segmental F1@50 increases from 51.36\% to 61.65\%, a gain of 10.29 percentage points. These aggregate results indicate that floorplan context benefits both window-level classification and temporal segmentation under the evaluated cross-environment protocol.

FCA-HAR improves direct-transfer accuracy by 10.73 percentage points, although its target-domain accuracy remains below its source-domain LOSO result. The datasets also differ in recording protocol and class distribution, so this gap cannot be attributed solely to furnishings. FCA could incorporate additional landmarks, such as kitchen counters or workstations, for broader activity taxonomies. This extension remains for future evaluation.

\section{Conclusion}
\label{sec:conclusion}

We propose a floorplan-context-aware strategy for cross-environment HAR using radar point clouds, requiring only a one-time registration of furniture landmarks in a new environment, with no target-domain activity annotations or model weight updates. Although the approach assumes that target activities exhibit informative spatial relationships with registered furniture landmarks, it offers an alternative to methods that rely exclusively on model adaptation for mitigating cross-environment performance degradation.

\clearpage
\bibliographystyle{IEEEbib}
\bibliography{refs}
\end{document}